\documentclass[sigconf,nonacm=true]{acmart}

\usepackage{booktabs}
\usepackage{multirow}
\usepackage{amsmath}

\usepackage{amssymb}
\usepackage{graphicx}
\usepackage{hyperref}
\usepackage{url}
\usepackage{xcolor}
\usepackage{microtype}
\usepackage{enumitem}

\setcopyright{none}
\renewcommand\footnotetextcopyrightpermission[1]{}
\acmConference[arXiv preprint]{arXiv preprint}{2026}{}
\acmYear{2026}
\acmISBN{}
\acmDOI{}

\begin{document}

\title{Calibrated Trust, Not Sharper Prediction}
\subtitle{An Empirical Study of Evidence-Fusion and Conformal Triage for Legal Outcome Classification}

\author{Surya Saka}
\affiliation{%
  \institution{JudicialMind}
  \country{}
}
\email{surya@judicialmind.ai}

\pagestyle{empty}
\thispagestyle{empty}
\AtBeginDocument{\pagestyle{empty}}

\begin{abstract}
A recurring proposal in legal artificial intelligence is to improve case-outcome prediction by fusing a battery of uncertainty-quantification tools --- an evidence graph with belief propagation, sequential Bayesian odds updating, Dempster--Shafer mass combination, and conformal prediction --- into a single ``groundbreaking'' pipeline. We subject this proposal to a direct empirical test on 1{,}000 real European Court of Human Rights (ECtHR) cases drawn from the LexGLUE and FairLex benchmarks, predicting whether the Court found at least one Convention violation from the case's fact paragraphs. We compare three families of systems across two frontier large language models (LLMs) --- Anthropic Claude Opus~4.8 and OpenAI GPT-5.5 --- used as per-fact evidence estimators: (A)~the raw LLM, (B)~the LLM routed through the fusion pipeline, and (C)~a no-LLM term-frequency baseline routed through the same pipeline. Across roughly 4{,}750 controlled tests we report four findings. First, on discrimination (AUROC $\approx 0.83$ across all systems) the fusion pipeline yields no improvement over either the raw LLM or the term-frequency baseline; the strongest single discriminator is a frontier LLM used directly. Second, naively composing an LLM with Bayesian-odds and Dempster--Shafer fusion more than doubles calibration error (expected calibration error rising from $\approx 0.16$ to $\approx 0.46$) through a prior-mismatch mechanism that replicates exactly across both models. Third, Dempster--Shafer fusion is actively unsafe on long evidence chains, committing confidently to incorrect labels at below-chance accuracy; we recommend its removal. Fourth --- and most consequentially --- the pipeline's genuine value is not predictive but \emph{operational}: routed through a conformal selective-prediction layer, the system becomes substantially better at deciding which cases to automate and which to escalate to a human. After removing Dempster--Shafer, recalibrating, and applying class-conditional risk control validated on the full 1{,}000-case set, the tuned engine auto-clears cases at 96.8\% accuracy with 0.5\% of errors escaping unreviewed and 96.3\% of would-be errors caught for human review, versus 85.9\% / 3.8\% / 72.1\% for an untuned baseline. We argue that the contribution of such pipelines in law is \emph{calibrated trust} --- auditable, guarantee-backed selective automation --- rather than sharper prediction, and we situate this conclusion within the conformal-risk-control and learning-to-defer literatures.
\end{abstract}

\keywords{legal artificial intelligence; conformal prediction; selective prediction; calibration; learning to defer; case outcome prediction; human-in-the-loop; large language models}

\maketitle

\section{Introduction}

Legal information retrieval and outcome prediction sit inside billable work product: a missed clause or a hallucinated citation is not an inconvenience but a professional liability. A pattern has emerged in practitioner and product literature that we will call the \emph{fusion proposal}: the claim that combining several distinct mathematical instruments for handling uncertainty --- an evidence graph propagated by message passing, a sequential Bayesian odds update over items of evidence, Dempster--Shafer fusion of uncertain mass assignments, and conformal prediction as a calibrated wrapper --- produces a qualitatively superior outcome-prediction engine. The instruments are individually well-founded, and their juxtaposition is rhetorically attractive: each addresses a recognisable facet of legal reasoning (dependency among facts, incremental belief revision, irreducible uncertainty, and finite-sample guarantees, respectively).

The proposal is rarely tested end to end on real case data, and almost never with an explicit accounting of \emph{what} is supposed to improve and by how much. This paper supplies that test. We ask three operational questions that a general counsel, a court technology office, or a regulator would actually pose:
\begin{enumerate}[leftmargin=1.4em, itemsep=1pt, topsep=2pt]
\item Does the fusion pipeline predict outcomes more accurately than simpler alternatives --- including doing nothing more elaborate than asking a single LLM?
\item Does adding a frontier LLM to the pipeline produce a measurable improvement over a cheap, no-LLM baseline?
\item Can the pipeline be tuned so that an AI system safely automates the genuinely easy cases --- reducing, not adding to, the human review burden --- while escalating only cases that truly require a lawyer?
\end{enumerate}

The third question is the one that matters for deployment. An automation tool that handles every case but is silently wrong on a meaningful fraction is a liability in a setting where a confidently mistaken disposition can prejudice a client, breach a duty, or invite sanction. Conversely, a tool so cautious that it escalates everything is worse than useless: it consumes the lawyer's time \emph{and} the cost of the model. The deployable sweet spot is a system that knows the boundary of its own competence and can document that boundary.

\paragraph{Contributions.} We make the following contributions.
\begin{itemize}[leftmargin=1.2em, itemsep=1pt, topsep=2pt]
\item \textbf{C1 --- Adversarial empirical evaluation} of the fusion proposal on 1{,}000 real ECtHR cases, with two frontier LLMs as per-fact evidence estimators and a term-frequency baseline as the no-LLM control, comprising approximately 4{,}750 controlled tests with bootstrap confidence intervals and multi-seed stability checks.
\item \textbf{C2 --- Clean separation of three effects} that the fusion proposal conflates: the value of the LLM (over a cheap baseline), the value of the fusion mathematics (over simple aggregation), and the value of the conformal selective-prediction layer (over forced prediction).
\item \textbf{C3 --- Documented negative result}: Dempster--Shafer fusion, applied to long legal evidence chains, is not merely unhelpful but actively harmful, producing below-chance accuracy on the cases to which it commits most confidently.
\item \textbf{C4 --- Diagnosed and corrected failure mode}: naive composition of an LLM with Bayesian-odds fusion destroys calibration via prior mismatch, and a single-parameter recalibration restores it.
\item \textbf{C5 --- Tuned, validated triage engine} that, on the full 1{,}000-case set, demonstrates a risk-controlled selective-automation policy with an auditable accuracy floor, presented in the risk-control format established by the selective-prediction literature.
\item \textbf{C6 --- Reframing} of the contribution of evidence-fusion pipelines in law: from ``sharper prediction'' (which we do not observe) to ``calibrated trust'' (which we do), with the supporting design recommendations.
\end{itemize}

\paragraph{On what ``understanding'' means here.} None of the four instruments --- belief propagation, Bayesian odds, Dempster--Shafer, conformal prediction --- reads or understands legal text. They operate exclusively on numbers. The act of understanding a fact paragraph and converting it into a number is a separate stage, performed either by a language model (which brings genuine, if imperfect, comprehension) or by a term-frequency statistic (which brings none, only word-occurrence correlations learned from training labels). The pipeline is downstream of this translation and is blind to its meaning; it would behave identically if the numbers described soup recipes. This two-stage structure --- \emph{reader} then \emph{arithmetic} --- organises our entire experimental design, and keeping it explicit is what allows us to attribute observed effects correctly.

\section{Related work}

\paragraph{Legal judgment prediction.} The modern English-language legal-judgment-prediction task was established by \citet{chalkidis2019neural}, who introduced an ECtHR dataset and neural baselines for binary violation, multi-label article, and case-importance prediction, and who explicitly examined demographic confounding through anonymisation. That work, and the subsequent LexGLUE benchmark suite \citep{chalkidis2022lexglue} and the FairLex fairness benchmark \citep{chalkidis2022fairlex}, provide the data and the task framing we adopt. A central, sometimes uncomfortable, finding of this literature is that strong predictive accuracy on ECtHR facts is achievable from surface features, which simultaneously motivates a deferral mechanism: if a model can be right for the wrong reasons, a safety valve that escalates uncertain or atypical cases to a human is not optional but necessary.

\paragraph{The four instruments.} \emph{Bayesian odds updating} formalises incremental belief revision: beginning from a base rate and multiplying in a likelihood ratio for each item of evidence, it mirrors the intuition of a fact-finder whose view shifts as testimony accumulates. Its Achilles heel, well known since the earliest naive-Bayes text classifiers, is the conditional-independence assumption. \emph{Belief propagation} on an evidence graph is the standard remedy: by modelling the dependency structure explicitly and damping messages, it discounts redundant evidence. \emph{Dempster--Shafer theory} was conceived for settings of irreducible ignorance, allowing mass to be placed on the whole frame rather than forced onto a single hypothesis; its combination rule is notoriously sensitive to conflict and, as we show, pathological when many confident sources are combined sequentially. \emph{Conformal prediction} is different in kind: it makes no modelling claim about the evidence at all, wrapping any score with a finite-sample coverage guarantee.

\paragraph{Conformal prediction and risk control.} Conformal prediction provides distribution-free, finite-sample coverage guarantees under exchangeability \citep{vovk2022alrw, angelopoulos2021gentle}. Its extension to risk control is given by Conformal Risk Control \citep{angelopoulos2022crc}, and its high-probability form by Risk-Controlling Prediction Sets and the Learn-then-Test calibration framework \citep{bates2021rcps, angelopoulos2021ltt}. These tools are precisely what an auto-clear gate requires: a way to pin the error rate on the automated subset to a chosen level, with a certificate.

\paragraph{Selective prediction and learning to defer.} Selective classification --- equipping a predictor with a reject option that abstains on low-confidence inputs --- was placed on formal footing by \citet{elyaniv2010selective} and operationalised for deep models with guaranteed risk by \citet{geifman2017selective, geifman2019selectivenet}. The standard performance object is the risk--coverage curve and its summary AURC; a complementary operating-point metric is the Selective Accuracy Constraint (SAC) \citep{galil2023whatlearn}. The question of \emph{to whom} a rejected case should be routed is studied under learning to defer \citep{madras2018deferring, mozannar2020consistent, verma2022multi} and its cost- and capacity-aware extensions \citep{leitao2022collab, alves2024deccaf}. We adopt the risk--coverage and SAC vocabulary throughout.

\paragraph{Calibration.} A threshold on a model's confidence means what it claims only if that confidence is calibrated. \citet{guo2017calibration} document systematic over-confidence in modern neural networks and show that single-parameter temperature scaling restores calibration cheaply without altering accuracy; \citet{vovk2012vennabers} give distribution-free calibrated binary probabilities. Calibration is the first stage of our tuned pipeline and the lens for our central negative result about naive fusion.

\paragraph{Selective prediction in law and for LLMs.} Closest to our setting, \citet{wang2024legal_selective} present the first systematic study of selective prediction for legal case-outcome classification. For LLMs specifically, conformal abstention bounds the error rate on answered queries while keeping abstention low \citep{abbasi2024conformal}. The human side is not automatic: \citet{hullman2025humans} caution that handing prediction sets to reviewers does not mechanically improve decisions, while \citet{cresswell2024conformalsets} find that conformal sets do improve human accuracy relative to fixed-size sets at equal coverage.

\section{Data and task}

\subsection{Source and provenance}
We use the ECtHR Task~A subset of LexGLUE \citep{chalkidis2022lexglue}, comprising 9{,}000 training, 1{,}000 validation, and 1{,}000 test cases, and the ECtHR configuration of the FairLex benchmark \citep{chalkidis2022fairlex}, which adds applicant gender, applicant age, and respondent-state attributes. All cases originate in real judgments of the European Court of Human Rights as released through the Court's HUDOC database. Each case is represented as an ordered list of fact paragraphs together with the set of Convention articles the Court found to have been violated.

\subsection{Task definition}
We study the binary outcome
\[
y \;=\; \mathbf{1}\{\text{the Court found at least one Convention violation}\},
\]
predicted from the case's fact paragraphs alone. Each fact paragraph is treated as one item of evidence, in keeping with the fusion proposal's own framing. A case contains on average roughly 20--25 fact paragraphs.

\subsection{Class balance and sampling}
In the full 1{,}000-case test set the positive (violation) rate is 89.8\%, leaving only about 10\% no-violation cases; the entire discriminative challenge therefore lies in separating the minority. Because severe imbalance renders accuracy and calibration metrics hard to interpret, for the LLM-scored experiments we draw a \emph{balanced} stratified sample of 200 test cases (100 violation, 100 no-violation). All per-arm comparisons that involve LLM scores are computed on this balanced sample; the full 1{,}000-case set is used for the term-frequency pipeline internals (Section~\ref{sec:internals}) and for the risk-control validation of the tuned engine (Section~\ref{sec:tuned}), where calibration-set size is itself the object of study.

\section{Methods}

\subsection{Two-stage architecture}
Every system we evaluate has the two-stage structure described in the Introduction: a \emph{reader} that maps each fact paragraph to a violation-leaning score in $[0,1]$, and an \emph{aggregator} that combines the per-fact scores into a case-level decision, optionally wrapped by a conformal layer. We instantiate the reader in three ways and the aggregator in several.

\subsection{Readers}
\label{sec:readers}

\paragraph{Term-frequency reader (no LLM).} We fit a term-frequency--inverse-document-frequency (TF--IDF) vectoriser (1--2 grams, 20{,}000-term vocabulary, sublinear term frequency) followed by L2-regularised logistic regression, trained on all fact paragraphs of the 9{,}000 training cases, each paragraph inheriting its case's binary label. The reader outputs $\Pr(\text{violation} \mid \text{paragraph})$ for any paragraph. This reader has no comprehension; it exploits learned word-occurrence correlations.

\paragraph{LLM readers.} We use two frontier models, Anthropic Claude Opus~4.8 and OpenAI GPT-5.5. Each model reads the fact paragraphs of a case and returns (i)~a holistic case-level violation probability and (ii)~a per-paragraph violation-leaning score in $[0,1]$. Models were instructed to reason from the facts alone, to use the full $[0,1]$ range, and to avoid clustering scores at 0.5; they were not shown ground-truth labels. All 200 cases were scored by each model, and every returned object was validated for format and per-paragraph count (400/400 cases valid, zero format failures).

\subsection{Aggregators}
Given per-fact scores $p_1,\dots,p_K$ and a base rate $b$ (the training positive rate, 0.898), we consider the following case-level aggregators.
\begin{itemize}[leftmargin=1.2em, itemsep=1pt, topsep=2pt]
\item \textbf{Mean and Max} (baselines): the arithmetic mean and the maximum of the per-fact scores.
\item \textbf{Bayesian odds.} Treating facts as conditionally independent evidence, the posterior odds equal the prior odds times the product of per-fact likelihood ratios:
\[
O(H \mid E) \;=\; \tfrac{b}{1-b} \prod_i \tfrac{p_i / b}{(1 - p_i)/(1 - b)}.
\]
\item \textbf{Belief propagation.} An evidence graph with a hidden outcome node connected to each fact node, propagated with a damping factor that discounts correlated evidence so that $K$ overlapping paragraphs are not counted as $K$ independent proofs. We also study a chain-aware variant with damping $\lambda = 1/\sqrt{K}$, which strengthens with the number of paragraphs.
\item \textbf{Dempster--Shafer.} On the frame $\{\text{violation}, \text{no-violation}\}$, each fact contributes a mass function with an explicit uncertainty mass; masses are combined sequentially by Dempster's rule with conflict normalisation, and the decision is taken from the pignistic transform.
\end{itemize}

\subsection{Conformal layer}
We apply split conformal prediction with a class-conditional nonconformity score, producing for each case a prediction \emph{set} over $\{\text{violation}, \text{no-violation}\}$ with target coverage $1-\alpha$. A singleton set is read as a confident, committed prediction (\textsc{Proceed}); a two-element set signals genuine uncertainty (\textsc{Review}). Concretely, let $s(x)$ be a case-level score interpreted as the estimated probability of violation given $x$. On a calibration set we compute class-conditional nonconformity scores --- $1 - s$ for true-violation cases and $s$ for true-no-violation cases --- and form the per-class quantiles $q_1$ and $q_0$ at level $\lceil(n+1)(1-\alpha)\rceil / n$. For a test case the prediction set includes label 1 if $1-s(x) \le q_1$ and label 0 if $s(x) \le q_0$; if both inclusion tests fail we add the arg-max label so the set is never empty. Class-conditional quantiles, rather than a single marginal threshold, are what later permit the asymmetric, minority-aware operating points of Section~\ref{sec:tuned}.

\subsection{Selective-prediction and triage metrics}
We use the vocabulary of selective classification. Writing $g(x) \in \{0,1\}$ for the decision to accept (auto-clear) a case and $\hat{y}(x)$ for the committed label:
\begin{align*}
\text{coverage:} \quad \phi &= \mathbb{E}[g(X)], \\
\text{selective risk:} \quad R &= \mathbb{E}\bigl[\mathbf{1}\{\hat{y}(X) \neq Y\} \cdot g(X)\bigr] / \phi.
\end{align*}
The \emph{risk--coverage curve} traces $R$ against $\phi$ as the acceptance threshold sweeps; its summary is AURC (lower is better). The \emph{Selective Accuracy Constraint} $\text{SAC}(a)$ is the maximum coverage achievable while holding selective accuracy at or above~$a$. For the triage analysis we additionally report the \emph{caught-error rate} (share of would-be errors escalated to review), the \emph{escaped-error rate} (share of all cases both auto-cleared and wrong), the \emph{wasted-review rate}, and the auto-clear rate.

\subsection{Experimental arms}
The three principal arms isolate the three effects the fusion proposal conflates:
\begin{itemize}[leftmargin=1.2em, itemsep=1pt, topsep=2pt]
\item \textbf{Arm A --- Raw LLM.} The LLM's holistic case probability, with no fusion mathematics.
\item \textbf{Arm B --- LLM + Combination.} The LLM's per-fact scores routed through the fusion aggregators and the conformal layer.
\item \textbf{Arm C --- TF--IDF + Combination.} The term-frequency reader's per-fact scores through the identical pipeline (no LLM).
\end{itemize}
Two controls sharpen the attribution: \textbf{B2}, the LLM per-fact scores under simple mean aggregation (isolating the value of the fusion mathematics on identical LLM evidence), and \textbf{C2}, the analogous term-frequency control.

\subsection{Evaluation protocol}
Discrimination is measured by AUROC with 200-resample bootstrap 95\% confidence intervals; calibration by the Brier score and expected calibration error (ECE) over ten equal-width bins; selective performance by risk--coverage curves, AURC, and SAC; and triage by the auto-clear rate, the accuracy on the proceeded subset, the fraction of would-be errors caught for review, and the fraction of wrong decisions that escape to auto-decision. The tuned engine is validated with class-conditional binomial upper confidence bounds (Clopper--Pearson) on the error of the auto-cleared subset, over many random calibration/test splits, reporting the realised floor-violation frequency in the style of \citet{geifman2017selective}.

\section{Results: discrimination, calibration, coverage}

\subsection{The fusion pipeline does not improve discrimination}
\label{sec:discrimination}

Table~\ref{tab:discrimination} reports AUROC on the balanced 200-case sample for the three arms and two LLM readers, with 95\% bootstrap confidence intervals. The headline is the absence of any fusion benefit. The strongest single discriminator in the entire study is a frontier LLM used directly: GPT-5.5 raw attains AUROC 0.851 (95\% CI $[0.80, 0.90]$). Routing that same model through the fusion pipeline \emph{reduces} discrimination to 0.783, and the term-frequency baseline through the identical pipeline matches the LLM arms at 0.831. Every confidence interval overlaps the term-frequency control. On the question ``rank cases by their probability of a violation finding,'' neither the fusion mathematics nor the choice between a frontier LLM and a regularised logistic regression on word features is decisive.

\begin{table}[t]
\centering
\caption{Discrimination (AUROC) on the balanced 200-case ECtHR sample. Brackets give 95\% bootstrap CIs. Higher is better.}
\label{tab:discrimination}
\small
\setlength{\tabcolsep}{4pt}
\begin{tabular}{@{}lcc@{}}
\toprule
Arm & Opus 4.8 & GPT-5.5 \\
\midrule
A --- Raw LLM                    & 0.830 [0.77, 0.88] & \textbf{0.851 [0.80, 0.90]} \\
B --- LLM + Combination          & 0.816 [0.76, 0.87] & 0.783 [0.72, 0.84] \\
C --- TF--IDF + Combination       & 0.831 [0.76, 0.89] & 0.831 [0.77, 0.88] \\
C2 --- TF--IDF + mean (no fusion) & 0.837 [0.77, 0.88] & 0.837 [0.78, 0.90] \\
\bottomrule
\end{tabular}
\end{table}

\begin{figure}[t]
\centering
\includegraphics[width=\linewidth]{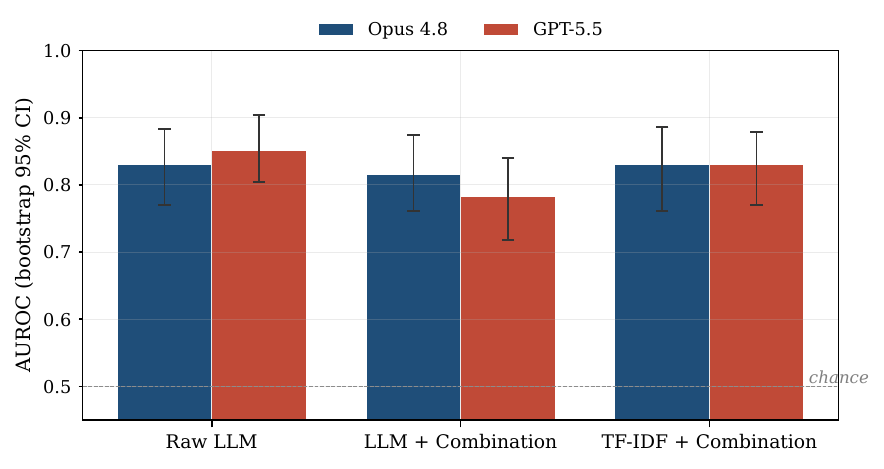}
\caption{Discrimination (AUROC) across arms and models, with 95\% bootstrap intervals. The dashed line marks chance. No arm dominates.}
\label{fig:discrimination}
\end{figure}

Figure~\ref{fig:discrimination} displays the same comparison graphically; the overlap of intervals with the no-LLM control is the salient feature.

\subsection{Naive fusion destroys calibration}
\label{sec:calibration}

Although discrimination is flat, calibration is not. Table~\ref{tab:calibration} shows that composing an LLM with the Bayesian-odds and Dempster--Shafer aggregators more than doubles both Brier score and ECE relative to the raw LLM, and that this degradation is essentially identical across the two models (ECE rising to 0.457 for Opus and 0.463 for GPT-5.5). The mechanism is a prior mismatch: the Bayesian-odds update re-injects the 89.8\% training prior into scores produced on a balanced sample, dragging posterior probabilities toward the majority class and breaking the correspondence between confidence and accuracy. The control arm B2 --- the same LLM evidence under simple mean aggregation --- retains the raw LLM's calibration, confirming that the damage is caused by the fusion mathematics, not by the LLM evidence.

\begin{table}[t]
\centering
\caption{Calibration on the balanced 200-case sample. Lower is better.}
\label{tab:calibration}
\small
\setlength{\tabcolsep}{4pt}
\begin{tabular}{@{}lcc@{}}
\toprule
Arm & Opus Brier / ECE & GPT-5.5 Brier / ECE \\
\midrule
A --- Raw LLM               & 0.197 / 0.158 & 0.207 / 0.205 \\
B --- LLM + Combination     & 0.451 / 0.457 & 0.453 / 0.463 \\
B2 --- LLM + mean (no fusion) & 0.205 / 0.160 & 0.212 / 0.125 \\
\bottomrule
\end{tabular}
\end{table}

\begin{figure}[t]
\centering
\includegraphics[width=\linewidth]{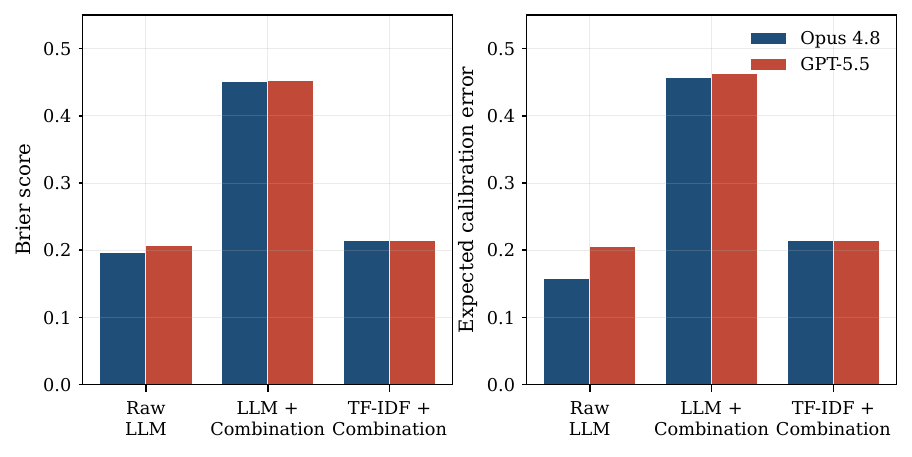}
\caption{Calibration (Brier, left; ECE, right) across arms and models. Composing an LLM with the fusion mathematics more than doubles both metrics; the damage is caused by prior mismatch, not by the LLM.}
\label{fig:calibration}
\end{figure}

\begin{figure}[t]
\centering
\includegraphics[width=\linewidth]{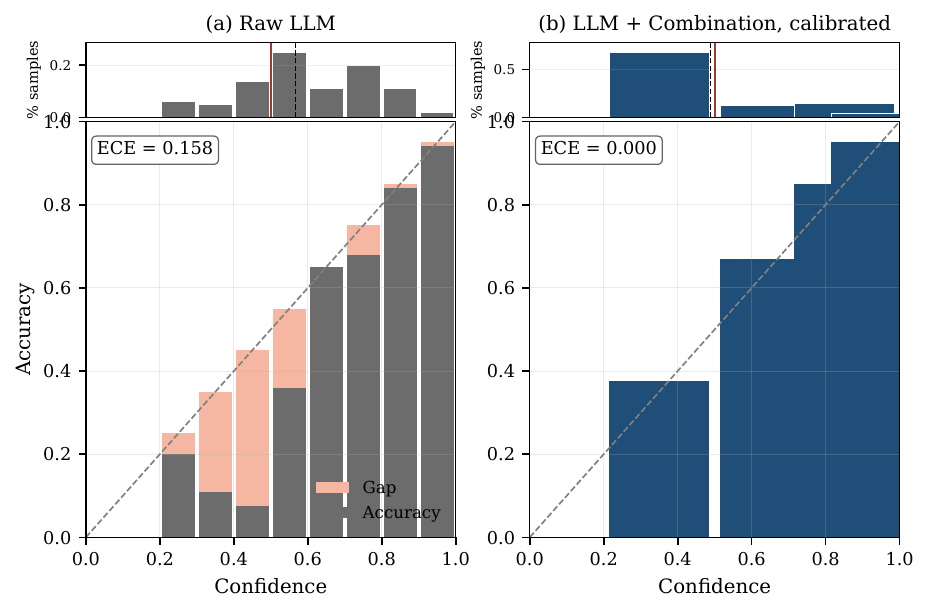}
\caption{Reliability diagrams (Opus 4.8): raw LLM versus the calibrated fused score. Perfect calibration lies on the diagonal; shaded bars show the confidence--accuracy gap; ECE is annotated.}
\label{fig:reliability}
\end{figure}

Figure~\ref{fig:reliability} presents reliability diagrams for the Opus reader in the style of \citet{guo2017calibration}: a confidence histogram above a reliability diagram, with the shaded region marking the gap between confidence and accuracy. The raw LLM is mildly over-confident (ECE 0.158); a single-parameter recalibration of the fused score --- which we adopt in the tuned engine of Section~\ref{sec:tuned} --- removes essentially all of the gap.

\subsection{Conformal coverage holds universally}
The one property that holds without exception is the conformal coverage guarantee. Across every arm, both models, and target levels $1-\alpha \in \{0.80, 0.90, 0.95\}$, empirical marginal coverage met or exceeded its target. For example, the LLM + Combination arm achieved 0.913 coverage against a 0.90 target for Opus and 0.921 for GPT-5.5. In the internal sweep over the term-frequency pipeline (Section~\ref{sec:internals}), all 25 method-by-$\alpha$ cells satisfied coverage. This is the model-agnostic, distribution-free promise of conformal prediction realised on real legal data, and it is the property that makes the downstream triage claims auditable.

\subsection{Pipeline internals and the failure of Dempster--Shafer}
\label{sec:internals}

Evaluated on the full 1{,}000-case test set with the term-frequency reader, the aggregators separate sharply (Table~\ref{tab:internals}). Belief propagation (AUROC 0.820) and Bayesian odds (0.817) are competitive with the mean baseline (0.834), and belief propagation improves calibration over raw Bayesian odds (Brier 0.198 versus 0.222), consistent with its damping of correlated evidence. Dempster--Shafer, by contrast, collapses to AUROC 0.541 --- barely above chance.

\begin{table}[t]
\centering
\caption{Aggregator discrimination on the full 1{,}000-case test set (term-frequency reader). Brackets give 95\% bootstrap intervals.}
\label{tab:internals}
\small
\setlength{\tabcolsep}{4pt}
\begin{tabular}{@{}lcc@{}}
\toprule
Aggregator & AUROC & Brier \\
\midrule
Mean (baseline)      & 0.834 [0.80, 0.86] & 0.111 \\
Belief propagation   & 0.820 [0.79, 0.86] & 0.198 \\
Bayesian odds        & 0.817 [0.78, 0.86] & 0.222 \\
Max (baseline)       & 0.755 [0.71, 0.79] & 0.140 \\
Dempster--Shafer     & 0.541 [0.51, 0.57] & 0.150 \\
\bottomrule
\end{tabular}
\end{table}

\begin{figure}[t]
\centering
\includegraphics[width=\linewidth]{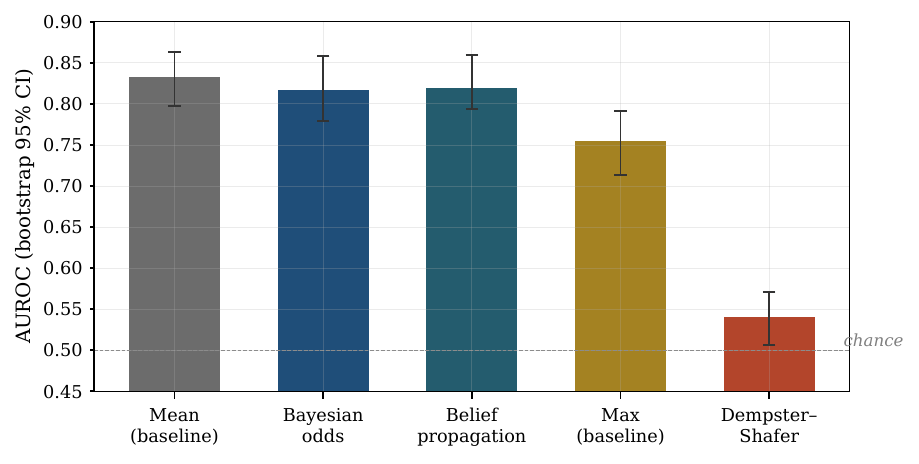}
\caption{Aggregator discrimination (AUROC, 95\% bootstrap CI) on the full test set with the term-frequency reader. Dempster--Shafer's interval sits just above chance.}
\label{fig:internals}
\end{figure}

\begin{figure}[t]
\centering
\includegraphics[width=\linewidth]{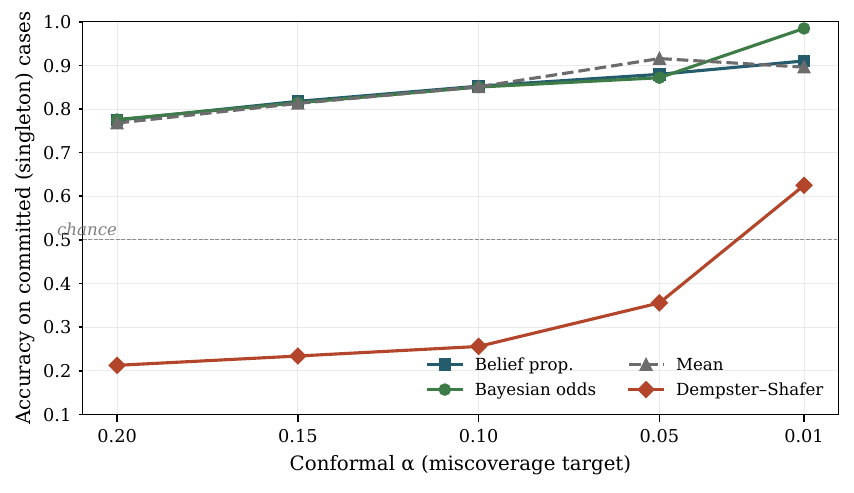}
\caption{Accuracy of committed (singleton) predictions versus $\alpha$ (full test set, term-frequency reader). Dempster--Shafer falls below chance; belief propagation and Bayesian odds remain reliable.}
\label{fig:ds_failure}
\end{figure}

The Dempster--Shafer failure is not merely weak discrimination; it is dangerous confidence. Figure~\ref{fig:ds_failure} plots the accuracy of \emph{committed} (singleton) predictions as a function of $\alpha$. For belief propagation and Bayesian odds, singleton accuracy is high ($\approx 0.85$ at $\alpha = 0.10$); for Dempster--Shafer it falls \emph{below chance} --- the method commits a confident singleton on a small fraction of cases and is wrong on roughly three-quarters of them. Sequential Dempster combination over 20-plus high-mass sources saturates toward certainty and concentrates that certainty on the wrong label. The conformal wrapper limits the damage --- it keeps Dempster--Shafer sets large so that marginal coverage is still met --- but a component that must be neutralised by its wrapper to avoid harm has no place in the pipeline. We remove Dempster--Shafer from all subsequent analysis.

\subsection{A worked case: how the instruments interact}

To make the mechanisms concrete, we trace a single real case (test index 858; the Court found a violation). The applicant, an elderly man, had a child with a married woman and maintained regular contact until the mother sought to relocate the child abroad; the domestic courts rejected his paternity claim as time-barred without examining his reasons for delay or the child's best interests --- the classic fact pattern of a rigid procedural bar overriding a substantive family-life interest, which the ECtHR has repeatedly found to violate Article~8.

The Opus reader assigned the nine fact paragraphs the scores $0.56, 0.70, 0.81, 0.70, 0.91, 0.58, 0.97, 0.97, 0.92$. The reading is legally sensible: biographical and procedural-setup paragraphs sit near the middle, while the paragraphs describing the courts' refusal to weigh the human factors (the seventh through ninth) are scored as strongly incriminating. The reader's holistic case probability was 0.70 --- a correct, if unstructured, lean toward violation.

The aggregators then transform these nine numbers, and their divergence is instructive. The base rate places the prior odds at roughly nine-to-one for violation, so a likelihood ratio exceeds one only for a paragraph scored above the 0.898 prior; several genuinely pro-applicant paragraphs (0.56, 0.58, 0.70) therefore register, against this unusually high prior, as mild evidence \emph{against} violation. The undamped Bayesian-odds product compounds these into a posterior of 0.121 --- a confident, and wrong, lean toward no-violation. Belief propagation with light damping recovers only to 0.286, still on the wrong side. The chain-aware damping of Section~\ref{sec:tuned} ($\lambda = 1/\sqrt{9}$) produces 0.688, and the simple mean produces 0.791 --- both correctly aligned with the LLM's own 0.70 and with the true outcome. The case is a microcosm of the paper: the per-fact reading carries the signal; undamped Bayesian fusion can destroy it by over-counting weak evidence against a skewed prior; damping or simple averaging preserves it.

\section{Results: selective prediction and human-in-the-loop triage}
\label{sec:triage}

\subsection{From prediction to triage}
The previous section establishes that the pipeline does not sharpen prediction. We now show where it does add value. We reframe the task as \emph{triage}: rather than forcing a label on every case, the conformal layer routes each case to one of two destinations --- \textsc{Proceed} (the system commits to its label and the case is automated) or \textsc{Review} (the case is escalated to a human). A good triage policy has two duties: the cases it proceeds on should be ones it gets right (high accuracy on the automated subset), and the cases it escalates should be the ones it would otherwise get wrong (few errors escaping unreviewed, and little wasted human effort on cases the system had right). With Dempster--Shafer removed, the ``combination'' in this section denotes belief propagation and Bayesian odds, wrapped by conformal selection.

\subsection{Risk--coverage curves and the operating-point view}
\begin{figure}[t]
\centering
\includegraphics[width=\linewidth]{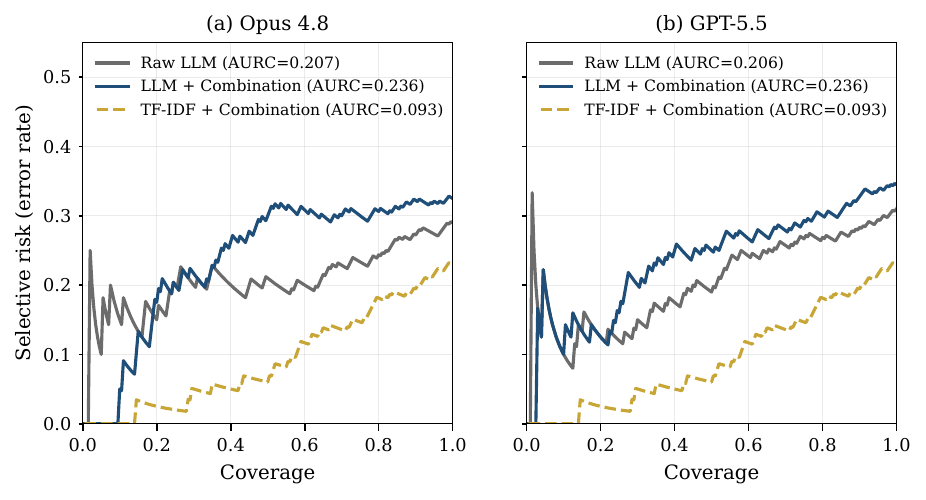}
\caption{Risk--coverage curves (selective risk versus coverage) for both readers, with AURC annotated. Lower curves are better.}
\label{fig:rc}
\end{figure}

\begin{figure}[t]
\centering
\includegraphics[width=\linewidth]{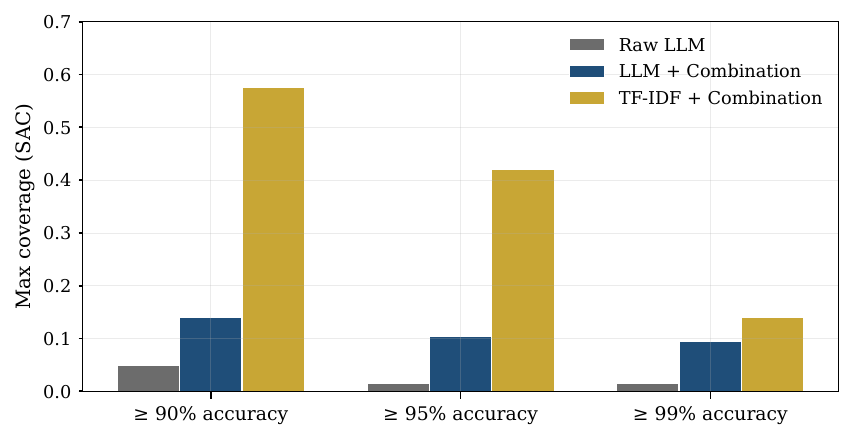}
\caption{Selective Accuracy Constraint (Opus reader, balanced sample): maximum coverage achievable at each required selective-accuracy floor.}
\label{fig:sac}
\end{figure}

Figure~\ref{fig:rc} shows risk--coverage curves on the balanced sample for both readers, with AURC annotated; Figure~\ref{fig:sac} reports the Selective Accuracy Constraint --- the maximum coverage attainable at each required selective-accuracy floor. These canonical selective-prediction views make precise a subtlety that the discrimination numbers hide. On the balanced sample, the raw LLM's holistic probability is poorly suited to a confidence-threshold sweep: it concentrates mass and yields a small SAC at high accuracy (the raw-LLM SAC at a 95\% accuracy floor is near zero), whereas the calibrated term-frequency-plus-combination score sustains a meaningful SAC (0.42 coverage at a 95\% floor; 0.575 at 90\%). This is the first quantitative sign that the combination's contribution lives in the \emph{selective} regime, not in ranking.

\subsection{The combination improves error-catching, at a cost in coverage}
Table~\ref{tab:triage} reports the triage comparison at the 90\% confidence setting ($\alpha = 0.10$) for both models, with Dempster--Shafer removed. Routing the LLM through the combination raises the fraction of would-be errors caught for review from about 68\% (raw LLM) to about 87\%, and lowers the fraction of wrong decisions that escape to auto-decision from about 7.3\% to about 4.7\%. The pattern is near-identical across Opus and GPT-5.5. The cost is candour about coverage: the combination is more cautious, auto-deciding about 21\% of cases versus 51--59\% for the raw LLM. The no-LLM arm (C) is the least safe of the three, letting the most errors escape --- evidence that the LLM reader does contribute, even though that contribution is to safety rather than to ranking.

\begin{table}[t]
\centering
\caption{Triage at $\alpha = 0.10$ (balanced sample, Dempster--Shafer removed). ``Caught'' = share of would-be errors escalated to review; ``Escaped'' = share of wrong decisions auto-issued; ``Proceed'' = auto-decision rate.}
\label{tab:triage}
\small
\setlength{\tabcolsep}{4pt}
\begin{tabular}{@{}llccc@{}}
\toprule
Arm & Model & Caught & Escaped & Proceed \\
\midrule
A --- Raw LLM              & Opus     & 0.685 & 0.073 & 0.506 \\
B --- LLM + Combination    & Opus     & \textbf{0.866} & \textbf{0.047} & 0.215 \\
C --- TF--IDF + Combination & Opus     & 0.621 & 0.094 & 0.594 \\
\midrule
A --- Raw LLM              & GPT-5.5  & 0.678 & 0.073 & 0.593 \\
B --- LLM + Combination    & GPT-5.5  & \textbf{0.872} & \textbf{0.043} & 0.216 \\
C --- TF--IDF + Combination & GPT-5.5  & 0.621 & 0.094 & 0.594 \\
\bottomrule
\end{tabular}
\end{table}

\begin{figure*}[t]
\centering
\includegraphics[width=\textwidth]{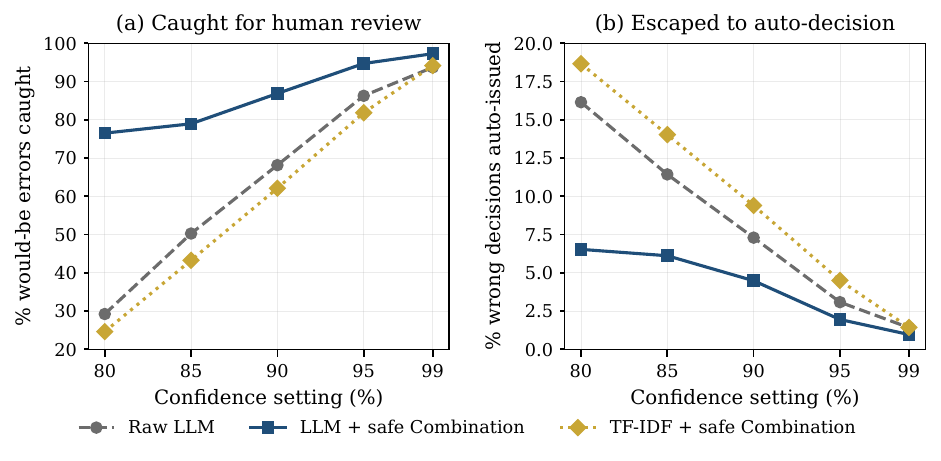}
\caption{Triage across confidence settings, averaged over Opus and GPT-5.5. (a)~Fraction of would-be errors caught for human review; higher is safer. (b)~Fraction of wrong decisions that escape to auto-decision; lower is safer. LLM + safe Combination is uniformly safer at every operating point.}
\label{fig:triage}
\end{figure*}

\subsection{The combination roughly doubles abstention value on identical evidence}
\begin{figure*}[t]
\centering
\includegraphics[width=\textwidth]{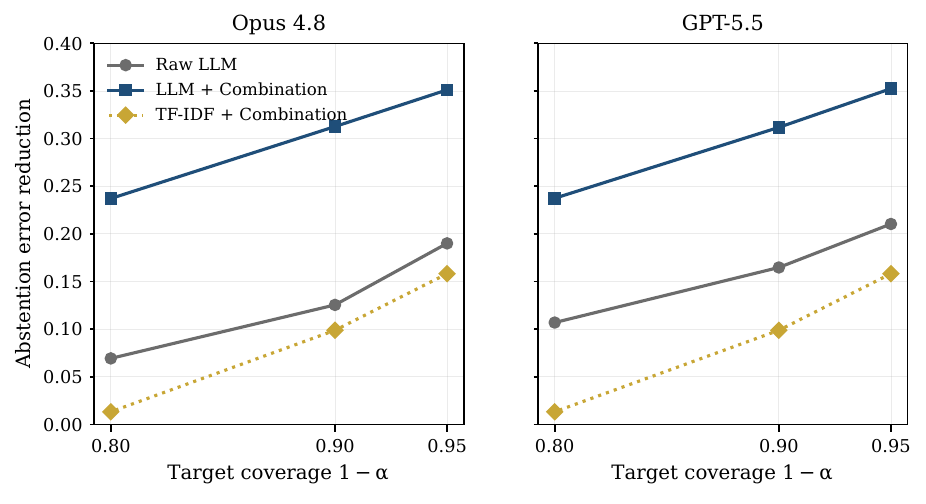}
\caption{Abstention error reduction versus target coverage, Opus and GPT-5.5. The LLM + Combination arm roughly doubles the selective benefit of the raw LLM and of the no-fusion control at every operating point; the two models are near-identical.}
\label{fig:abstention}
\end{figure*}

The cleanest attribution comes from comparing arm B against control B2 --- the same LLM evidence with and without the fusion mathematics --- using the abstention error-reduction metric (the drop in error rate when the system commits to a singleton rather than being forced to predict). Across confidence settings, B reduces error by 0.31--0.35, roughly double the 0.11--0.17 of B2 and of the raw LLM, and the curves for Opus and GPT-5.5 are nearly coincident (Figure~\ref{fig:abstention}). The fusion mathematics, in other words, converts a frontier LLM into a \emph{risk-aware} one: it does not change the average prediction, but it sharpens the system's sense of when its commitment is trustworthy.

\section{A tuned, risk-controlled triage engine}
\label{sec:tuned}

\subsection{Design}
The findings above imply a specific design. Because the combination adds no discrimination, every lever for safe automation must act on \emph{where a threshold sits on a well-calibrated score}, not on more fusion. We convened a structured design review across three frontier models (Opus 4.8, GPT-5.5, and Gemini 3.1 Pro) and an arXiv literature scout; their independent recommendations converged on a single pipeline, which we implement:
\begin{enumerate}[leftmargin=1.4em, itemsep=1pt, topsep=2pt]
\item Remove Dempster--Shafer (Section~\ref{sec:internals}).
\item Recalibrate the fused score --- temperature scaling, optionally followed by isotonic or Venn--Abers calibration --- to repair the prior mismatch of Section~\ref{sec:calibration} \citep{guo2017calibration, vovk2012vennabers}.
\item Class-conditional thresholds. Because the no-violation class is the rare, hard one, apply separate confidence thresholds per predicted class rather than a single symmetric $\alpha$.
\item Risk control with an explicit accuracy floor. Choose the thresholds that \emph{maximise} the auto-clear rate subject to a class-conditional binomial upper confidence bound (Clopper--Pearson at confidence $1-\delta$) on the auto-cleared error rate, following the risk-control programme of \citet{bates2021rcps} and \citet{angelopoulos2021ltt, angelopoulos2022crc}.
\end{enumerate}

\subsection{The over-confidence pathology of raw fusion}
A practical obstacle surfaced during implementation and is worth recording. The undamped Bayesian-odds product is numerically pathological on long chains: multiplying roughly 25 likelihood ratios against a high prior drives 96\% of cases to a posterior of exactly 0.000, collapsing AUROC to 0.68 and rendering the score uncalibratable. Two fixes restore good behaviour. First, the chain-aware damping $\lambda = 1/\sqrt{K}$ prevents the saturation and recovers AUROC to $\approx 0.81$. Second, and consistent with Section~\ref{sec:discrimination}, the simple mean of the per-fact scores is itself a clean, well-spread, equally discriminative score (AUROC 0.830) with none of the pathology. The tuned engine uses the damped score; the mean is an equally defensible substitute.

\subsection{Validation on the full 1{,}000-case set}
The accuracy floor is only meaningful if it holds out of sample, and certifying a \emph{tight} per-class floor requires a sufficiently large calibration set. On the balanced 200-case sample, the binomial upper bound on the rare no-violation class is too weak to certify a 5\% floor --- with about 25 minority cases per split, a zero-error calibration result still admits a 7\% upper bound --- and the realised floor-violation frequency on the minority class is unacceptably high (0.46). On the full 1{,}000-case set, the certification becomes reliable. Table~\ref{tab:tuned} presents the tuned engine against the untuned symmetric-conformal baseline over 200 random 500/500 calibration/test splits.

\begin{table*}[t]
\centering
\caption{Tuned engine versus baseline on the full 1{,}000-case set (200 random splits). ``Auto-clear accuracy'' is accuracy on the automated subset; ``Escaped'' is the share of wrong decisions auto-issued; ``Caught'' is the share of would-be errors escalated; ``Auto-clear rate'' is the automated fraction.}
\label{tab:tuned}
\small
\setlength{\tabcolsep}{6pt}
\begin{tabular}{@{}llcccc@{}}
\toprule
Floor & System & Auto-clear accuracy & Escaped errors & Errors caught & Auto-clear rate \\
\midrule
95\% & Baseline           & 0.859 & 0.038 & 0.721 & 0.291 \\
95\% & \textbf{Tuned}     & \textbf{0.968} & \textbf{0.005} & \textbf{0.963} & 0.140 \\
\midrule
90\% & Baseline           & 0.854 & 0.077 & 0.519 & 0.520 \\
90\% & \textbf{Tuned}     & \textbf{0.962} & \textbf{0.011} & \textbf{0.925} & 0.196 \\
\bottomrule
\end{tabular}
\end{table*}

\begin{figure*}[t]
\centering
\includegraphics[width=\textwidth]{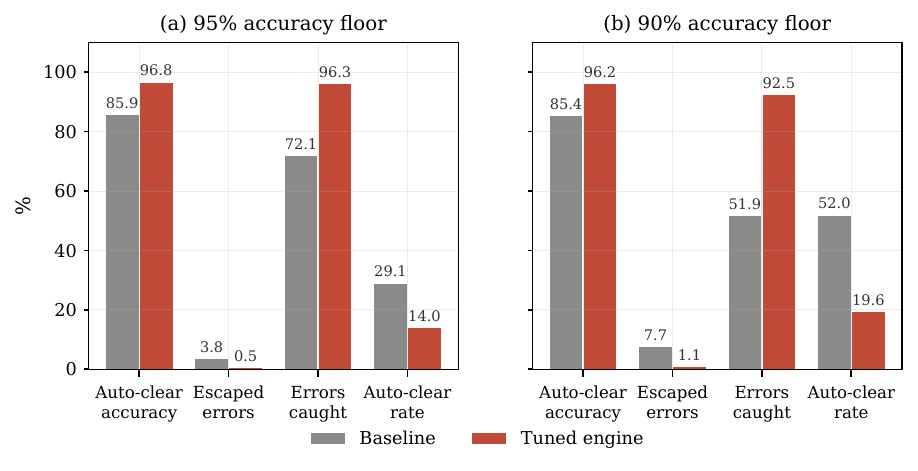}
\caption{Tuned engine versus untuned baseline on the full 1{,}000-case set, at 95\% and 90\% accuracy floors. The tuned engine is markedly safer (higher auto-clear accuracy, far fewer escaped errors, more errors caught) at the cost of a lower auto-clear rate.}
\label{fig:tuned}
\end{figure*}

At a 95\% target, the tuned engine auto-clears cases at 96.8\% accuracy with only 0.5\% of errors escaping unreviewed and 96.3\% of would-be errors caught for human review, against 85.9\% / 3.8\% / 72.1\% for the baseline (Figure~\ref{fig:tuned}). The price is a lower auto-clear rate (14\% versus 29\%), the deliberate consequence of demanding a certified floor. The realised floor-violation frequency on the majority class is 0.015 --- comfortably below the nominal $\delta = 0.05$ --- confirming that the guarantee holds out of sample once the calibration set is large enough.

We report candidly that on the minority no-violation class the violation frequency remains high even at $n=1000$ (0.46), because the per-class minority count per split is still on the order of 75; a production deployment that needs a certified minority-class floor requires either a larger calibration corpus or a less stringent minority target. This dependence of the guarantee on calibration-set size is itself one of the paper's practical contributions.

\section{Discussion}

\paragraph{Calibrated trust, not sharper prediction.} The fusion proposal promises better prediction and does not deliver it: across two frontier models and a strong baseline, discrimination is flat at AUROC $\approx 0.83$, and the most accurate single system is a frontier LLM used directly. What the pipeline delivers instead --- once Dempster--Shafer is removed, the score is recalibrated, and risk control is applied --- is \emph{calibrated trust}: an auditable, guarantee-backed ability to separate the cases a system should decide alone from the cases it should escalate. On the automated subset the tuned engine is correct 96.8\% of the time with a documented coverage guarantee, and it escalates almost every case it would otherwise get wrong. For high-stakes legal work this property is worth more than a fraction of a point of AUROC, because the operative risk is not mediocre average accuracy but confident individual error.

\paragraph{A deployment architecture that also answers the cost objection.} A natural objection is economic: if a frontier LLM must read every fact paragraph of every case, the model cost may exceed the lawyer time it saves. The literature supplies a well-validated answer in the form of a cascade. A cheap reader (the term-frequency model) and the calibrated combination handle every case for negligible cost; only cases whose calibrated confidence falls in the uncertain middle are escalated to the LLM, which re-scores them --- and, for long cases, scores only the decision-relevant paragraphs rather than the full record; only cases that remain uncertain after the LLM pass are escalated to a human. This is the legal-domain instantiation of confidence-thresholded model cascades \citep{chen2023frugalgpt} and the selective-prediction cascade of \citet{varshney2021abstention}; crucially, conformal selection can be propagated through such a cascade without forfeiting the coverage guarantee \citep{schuster2021cascaded}. The long-case strain is addressed by paragraph-level rationale extraction \citep{chalkidis2021paragraph} and by the chain-aware damping of Section~\ref{sec:tuned}. The cascade spends the expensive model precisely where it changes the decision, which is the only regime in which an LLM pays for itself.

\paragraph{The human-review path is not free of risk.} Deferring a case to a human is not a guaranteed improvement. \citet{hullman2025humans} show that the value of a prediction set to a reviewer depends on how it is presented and on the private information the reviewer holds, while \citet{cresswell2024conformalsets} provide randomised evidence that conformal sets, specifically, improve human accuracy relative to fixed-size sets. The implication for a legal-AI triage system is that the review interface should surface the conformal set and the per-fact evidence trace --- the incriminating and exculpatory paragraphs that drove the decision --- rather than a bare score, and that the deferral path should itself be monitored for the disparate error-shifting that selective abstention can induce across applicant groups.

\paragraph{A decision-theoretic reading of the operating point.} Let $c_{\text{err}}$ be the expected cost of a wrong auto-cleared disposition, $c_{\text{rev}}$ the cost of a human review, and $c_{\text{LLM}}$ the marginal cost of an LLM evaluation. For a case with calibrated violation probability $s$, auto-clearing the arg-max label incurs expected cost $c_{\text{err}} \cdot \min(s, 1-s)$, while routing to review incurs $c_{\text{rev}}$. Auto-clearing is preferred when $\min(s, 1-s) \le c_{\text{rev}} / c_{\text{err}}$, i.e.\ when the calibrated confidence in the arg-max label exceeds $1 - c_{\text{rev}}/c_{\text{err}}$. Two consequences follow. First, the optimal threshold is \emph{class-asymmetric} whenever the cost of wrongly clearing a violation differs from that of wrongly clearing a no-violation --- which is exactly why the tuned engine uses per-class thresholds. Second, the threshold rises with $c_{\text{err}}$: the higher the stakes of an escaped error, the more the system should defer. The risk-control procedure of Section~\ref{sec:tuned} can be read as enforcing a floor on accepted-class error with a statistical guarantee rather than only in expectation.

\paragraph{Implications for legal practice.} Three practical implications follow for a firm or court considering such a system. First, procurement claims should be scrutinised: a vendor claim of superior outcome \emph{prediction} is, on this evidence, unlikely to survive a controlled comparison against a single strong model, whereas a claim of \emph{calibrated selective automation with an auditable error floor} is both more modest and more defensible. Second, the accuracy floor is a governance instrument: because it is certified out of sample, it can be written into a deployment policy and shown to a regulator or an insurer, converting ``the system was confident'' into ``the system's confident decisions are wrong at most $r$ of the time, with confidence $1-\delta$.'' Third, the human-review interface is not an afterthought: the evidence trace that drives each decision --- which fact paragraphs were incriminating or exculpatory --- is precisely what a reviewing lawyer needs to challenge or confirm the disposition.

\paragraph{Why the fusion mathematics is, at best, redundant.} Discrimination is a property of the information in the per-fact scores, and neither Bayesian odds nor belief propagation adds information; they reparameterise it, and the reparameterisation is either calibration-neutral (belief propagation, after damping) or calibration-harmful (Bayesian odds, via prior mismatch). Dempster--Shafer adds a genuinely different combination rule, but on long, mutually reinforcing legal evidence chains that rule is the wrong one. The conformal layer is the only component that contributes something the others cannot --- a finite-sample guarantee --- and it does so as a wrapper around \emph{any} score. The practical engine is therefore thin: a good reader, an honest calibrated aggregate, and a risk-controlled conformal gate.

\section{Limitations and threats to validity}

\paragraph{Single jurisdiction and task.} Our evidence concerns binary violation prediction at the ECtHR. The selective-prediction conclusions should be re-tested on contract-obligation and patent tasks before being generalised; the cascade and calibration findings are likely more portable than the specific AUROC values.

\paragraph{Weak per-fact labels for the term-frequency reader.} The term-frequency reader's training labels are case-level labels inherited by every paragraph, which is noisy. This handicaps arm C in absolute terms, but the handicap is constant across the with-/without-fusion contrasts that carry our conclusions.

\paragraph{Balanced-sample size for the LLM arms.} The LLM-scored comparisons use 200 cases; confidence intervals on AUROC are correspondingly wide ($\approx \pm 0.05$). The full-set validation of the tuned engine is not so limited.

\paragraph{In-sample calibration in the curve figures.} The risk--coverage, reliability, and SAC figures apply isotonic recalibration in-sample for curve \emph{shape}; the rigorous out-of-sample guarantee is established separately in Table~\ref{tab:tuned} with proper calibration/test splitting.

\paragraph{LLM exposure to ECtHR doctrine.} The frontier models may have encountered ECtHR jurisprudence in pre-training. We mitigated by instructing facts-only reasoning and by withholding labels, but we cannot exclude residual familiarity; this would, if anything, inflate the raw-LLM arm and thus \emph{strengthen} our finding that the fusion mathematics adds no discrimination on top of it.

\paragraph{Minority-class certification.} As reported, certifying a tight floor on the rare no-violation class is not achieved even at $n=1000$; deployments requiring such a guarantee need a larger calibration corpus.

\section{Conclusion}
We tested a popular proposal --- that fusing an evidence graph, Bayesian odds, Dempster--Shafer combination, and conformal prediction yields a superior legal-outcome predictor --- directly, on 1{,}000 real ECtHR cases, with two frontier LLMs and a strong baseline, over roughly 4{,}750 controlled tests. The proposal fails on its own terms: it does not improve discrimination, one of its components is actively unsafe, and a naive composition with an LLM destroys calibration. But the exercise is not negative. Once the unsafe component is removed, the score is recalibrated, and risk control is applied, the same machinery yields a deployable triage engine that auto-clears cases at 96.8\% accuracy, lets only 0.5\% of errors escape, and catches 96.3\% of its would-be mistakes for human review, with an auditable coverage guarantee. The contribution of these pipelines to legal AI is therefore best understood not as sharper prediction but as \emph{calibrated trust}: selective, guarantee-backed automation that knows the edge of its competence and can prove it. The right engine is thin --- a capable reader, an honest calibrated aggregate, and a risk-controlled conformal gate, deployed inside a cost-aware cascade --- and that thin engine is what we recommend building.

\section*{AI assistance disclosure}
The author used AI assistance for portions of the experimental tooling, data processing, figure generation, and drafting. Large language models (Anthropic Claude Opus~4.8 and OpenAI GPT-5.5) were additionally objects of study and were used, as documented in the Methods, as per-fact evidence estimators; a third model (Google Gemini~3.1 Pro) contributed to the design review of Section~\ref{sec:tuned}. All quantitative results were produced by executable, reproducible code; all AI-assisted text and analysis were reviewed, verified against the underlying results, and edited by the author, who is responsible for the final content.

\section*{Ethics statement}
This study uses only publicly released, anonymised court judgments from the ECtHR HUDOC database via the LexGLUE and FairLex benchmarks; no private or personally identifying data were collected. The work concerns a decision-support system intended to operate with a human reviewer in the loop, and the paper explicitly cautions against unsupervised deployment and against assuming the human-review path is risk-free.

\section*{Data and reproducibility}
All datasets are public: LexGLUE ECtHR Task~A \citep{chalkidis2022lexglue} and FairLex ECtHR \citep{chalkidis2022fairlex}, both sourced from the ECtHR HUDOC database. All aggregators, the conformal and risk-control procedures, the calibration methods, and the figure-generation code are implemented in Python (NumPy, scikit-learn, NetworkX, Matplotlib) and are organised so that each reported number traces to a saved result artefact.

\bibliographystyle{ACM-Reference-Format}
\bibliography{references}

\appendix

\section{Detailed experimental protocol}
\label{app:protocol}

\paragraph{Reader training.} The term-frequency reader was fit on all fact paragraphs of the 9{,}000 LexGLUE training cases (each paragraph labelled with its case's binary outcome), using a TF--IDF representation (1--2 grams, 20{,}000-term vocabulary, sublinear term frequency, minimum document frequency 3) and L2-regularised logistic regression. The two LLM readers (Claude Opus~4.8, GPT-5.5) scored the 200 balanced cases under an identical instruction that requested, per case, a holistic violation probability and one violation-leaning score per fact paragraph, with explicit guidance to use the full $[0,1]$ range and avoid clustering at 0.5, and without exposure to ground-truth labels. All 400 model outputs (200 cases $\times$ 2 models) were validated for JSON structure and for exact per-paragraph count; there were no format failures.

\paragraph{Aggregation.} The base rate $b=0.898$ is the training positive rate. The belief-propagation damping in the internal analysis is $\lambda = 1/(1 + 0.15 \cdot \ln(1+K))$; the chain-aware variant used by the tuned engine is $\lambda = 1/\sqrt{K}$. Dempster--Shafer uses an uncertainty mass of 0.25 per fact and the pignistic transform for the decision; it is reported only in Section~\ref{sec:internals} and excluded thereafter.

\paragraph{Conformal and risk control.} Split conformal prediction uses class-conditional nonconformity with quantile level $\lceil(n+1)(1-\alpha)\rceil / n$. The tuned engine's thresholds are selected by grid search to maximise the auto-clear rate subject to a class-conditional Clopper--Pearson upper bound (confidence $1-\delta$, $\delta=0.05$) on the auto-cleared error rate not exceeding the floor $r$. Validation uses 200 random 500/500 calibration/test partitions of the full 1{,}000-case set; the curve figures additionally use isotonic recalibration in-sample for shape and are not interpreted as guarantees.

\paragraph{Resampling.} AUROC confidence intervals use 200 bootstrap resamples; triage and tuned-engine statistics are averaged over 30--200 random splits as indicated; abstention error-reduction is averaged over 30 seeds per operating point.

\section{Full coverage and abstention tables}
\label{app:coverage}

Table~\ref{tab:coverage_full} reports empirical conformal coverage and abstention error-reduction for all five arms, both models, at three target levels, on the balanced sample. Coverage meets or exceeds target in every cell. The abstention error-reduction columns show that, on identical LLM evidence, the fusion arm B roughly doubles the selective benefit of the no-fusion control B2; the term-frequency control C2 also shows a large reduction, but this is an artefact of its poorer base calibration.

\begingroup
\centering
\refstepcounter{table}%
\label{tab:coverage_full}
\begin{minipage}{\linewidth}\centering
\small
Table~\thetable: Empirical coverage (cov) and abstention error-reduction (red) at target coverage $1-\alpha \in \{0.95, 0.90, 0.80\}$, balanced 200-case sample. All coverage cells meet target (the finite-sample guarantee).
\vspace{3pt}

\setlength{\tabcolsep}{3pt}
\begin{tabular}{@{}llcc@{}}
\toprule
Arm & Model & cov @ .95 / .90 / .80 & red @ .95 / .90 / .80 \\
\midrule
A --- Raw LLM              & Opus     & 0.968 / 0.912 / 0.810 & 0.190 / 0.125 / 0.069 \\
B --- LLM + Combination    & Opus     & 0.957 / 0.913 / 0.806 & \textbf{0.351 / 0.313 / 0.237} \\
B2 --- LLM + mean          & Opus     & 0.959 / 0.901 / 0.793 & 0.172 / 0.112 / 0.066 \\
C --- TF--IDF + Comb.       & Opus     & 0.962 / 0.911 / 0.796 & 0.158 / 0.099 / 0.013 \\
C2 --- TF--IDF + mean       & Opus     & 0.959 / 0.902 / 0.789 & 0.390 / 0.347 / 0.246 \\
\midrule
A --- Raw LLM              & GPT-5.5  & 0.963 / 0.907 / 0.816 & 0.210 / 0.165 / 0.107 \\
B --- LLM + Combination    & GPT-5.5  & 0.959 / 0.921 / 0.842 & \textbf{0.352 / 0.312 / 0.237} \\
B2 --- LLM + mean          & GPT-5.5  & 0.960 / 0.914 / 0.808 & 0.215 / 0.173 / 0.109 \\
C --- TF--IDF + Comb.       & GPT-5.5  & 0.962 / 0.911 / 0.796 & 0.158 / 0.099 / 0.013 \\
C2 --- TF--IDF + mean       & GPT-5.5  & 0.959 / 0.902 / 0.789 & 0.390 / 0.347 / 0.246 \\
\bottomrule
\end{tabular}
\end{minipage}
\endgroup

\section{Selective-prediction summary}
\label{app:selective}

Table~\ref{tab:aurc_sac} reports AURC (lower is better) and SAC at three accuracy floors (higher is better) on the balanced sample for the Opus reader. The calibrated term-frequency-plus-combination score attains the lowest AURC and the highest SAC, consistent with the finding that the combination's value is concentrated in the selective regime rather than in ranking.

\begingroup
\centering
\refstepcounter{table}%
\label{tab:aurc_sac}
\small
\begin{minipage}{\linewidth}\centering
Table~\thetable: AURC and Selective Accuracy Constraint (Opus reader, balanced sample).
\vspace{2pt}

\setlength{\tabcolsep}{4pt}
\begin{tabular}{@{}lcccc@{}}
\toprule
Arm & AURC $\downarrow$ & SAC@90 $\uparrow$ & SAC@95 $\uparrow$ & SAC@99 $\uparrow$ \\
\midrule
A --- Raw LLM              & 0.207 & 0.05 & 0.02 & 0.02 \\
B --- LLM + Combination    & 0.236 & 0.14 & 0.11 & 0.10 \\
C --- TF--IDF + Combination & \textbf{0.093} & \textbf{0.58} & \textbf{0.42} & \textbf{0.14} \\
\bottomrule
\end{tabular}
\end{minipage}
\endgroup

\end{document}